\documentclass{article} 
\usepackage{iclr2027_conference,times}

\usepackage{amsmath,amsfonts,bm}

\def\eqref#1{equation~\ref{#1}}

\def\1{\bm{1}}

\DeclareMathAlphabet{\mathsfit}{\encodingdefault}{\sfdefault}{m}{sl}
\SetMathAlphabet{\mathsfit}{bold}{\encodingdefault}{\sfdefault}{bx}{n}

\usepackage{hyperref}
\usepackage{url}
\usepackage{amsmath,amssymb}
\usepackage{booktabs}
\usepackage{multirow}
\usepackage{graphicx}
\usepackage{xcolor}
\usepackage{tikz}
\usepackage{pgfplots}
\pgfplotsset{compat=1.17}
\usetikzlibrary{positioning,arrows.meta,decorations.pathreplacing,calc,patterns}
\usepackage{enumitem}
\usepackage{needspace}
\setlist{nosep,leftmargin=*}

\newcommand{\polar}{PoLar}
\newcommand{\cs}{\textsc{ConfSwitch}}
\newcommand{\mrg}{\textsc{Margin}}
\newcommand{\pp}{\textsc{Program}}
\newcommand{\dir}{\textsc{Direct}}
\newcommand{\cotarm}{\textsc{CoT}}
\newcommand{\orc}{\textsc{Oracle}}
\newcommand{\torc}{\textsc{Transferable Oracle}}
\newcommand{\vote}{\textsc{Vote}}
\newcommand{\best}[1]{\textbf{#1}}
\newcommand{\second}[1]{\underline{#1}}
\newcommand{\finding}[1]{\textbf{Finding #1.}}

\title{Dynamic Routing as a New Dimension for Test-time Versatility of LLMs}

\author{Michal Štefánik$^{\clubsuit\spadesuit}$\ \ \ \   Marek Kadlčík$^\clubsuit$\ \ \   Josef Kuchař$^\clubsuit$\ \ \   Michal Spiegel$^{\clubsuit}$\vspace{6pt}\\
    \normalsize{$^\spadesuit$R\&D Centre for Large Language Models, National Institute of Informatics, Japan} \\
  \normalsize{$^\clubsuit$TransformersClub, Faculty of Informatics, Masaryk University}
 }

\iclrfinalcopy 
\begin{document}

\maketitle
\lhead{Preprint}

\begin{abstract}
Beyond scaling their parameters and data, large language models currently gain versatility on new problems along a single axis: the tokens they spend on chain-of-thought (\cotarm).
We investigate whether dynamic routing programs, which execute a subset of the model's layers or iterate some of them, can open a second axis of test-time adaptation, complementary to \cotarm\ and free of any gradient update.
Prior work showed that such programs exist and bring accuracy and efficiency gains on problems similar to those they were trained on; we ask whether they can also be identified rapidly, from a handful of demonstrations ($n \in \{3, 10\}$), by a strategy that transfers across models and tasks without training.
First, we find that strategies that select programs by the probability they assign to the demonstrations' labels, arbitrated by the model's own confidence, bring consistent gains: on average over the 49 tasks of MMLU and substantially on four of seven models, and most of all on far out-of-distribution tasks such as ARC-AGI, where programs double the accuracy of a 7B model whose \cotarm\ fails.
Second, on MMLU across the seven post-trained models, routing complements \cotarm\ in practice: the two succeed on different queries, and their composition exceeds \cotarm\ alone.
Despite these gains, our analyses show that confidence-based selection leaves much of the potential untapped, in two places in particular: (1) in surfacing the routing potential that is already present early in pre-training but becomes harder to select after post-training, and (2) in making models robust to the refinements routing introduces, since unsuccessful routes tend to drive the residual stream out of the distribution that the following layers expect.
Together, our results point to dynamic routing as a paradigm for extending the plasticity of existing and future LLMs in rapid test-time adaptation.
\end{abstract}

\section{Introduction}
\label{sec:intro}

Most capabilities of large language models arrive with scale, of the parameters, of the data, or, most recently, of the inference-time computation spent on chain-of-thought~\citep{wei2022cot,snell2025scaling,deepseek2025r1}. The last is the one axis along which a deployed model adapts to a new problem: given a harder question, it generates a longer chain. That axis is expensive, every token being a sequential step, and not universal, since on some problems a longer chain makes the answer worse (\S\ref{sec:arc}). We ask whether a fixed set of parameters supports a \emph{second} axis of test-time adaptation, one that changes how the model computes rather than how long it generates, exploitable without changing the parameters, without a training set, and within minutes of meeting a new task.

Interpretability work describes the forward pass as a \emph{gradual refinement} of a prediction largely formed early~\citep{stefanik2026lmslearn,belrose2023tuned,jastrzebski2018residual}: consecutive layers apply small, similar updates and middle layers are largely interchangeable and removable~\citep{sun2024painters,gromov2025unreasonable,men2024shortgpt,csordas2025depth}, which makes the order and multiplicity in which layers are applied a free variable at inference time. Looped transformers exploit this by design but train a new model~\citep{dehghani2019universal,giannou2023looped,geiping2025scaling}. At the other end, \citet{li2025cola,li2026polar} showed that an \emph{existing} fixed-depth model already admits many alternative execution orders: a \emph{Program of Layers} (\polar) skips or repeats blocks of pretrained layers per input, and an offline search finds, for most inputs, programs that correct the full pass's errors. This reveals a substantial \emph{potential} for dynamic routing, but leaves open how to harness it: the predictor of \citet{li2026polar} is trained and evaluated on one distribution of mathematical problems, the closest trained router~\citep{heakl2025drllm} loses accuracy on unseen tasks (\S\ref{sec:related}), and routing has never been composed with chain-of-thought.

This work investigates dynamic routing as that second axis under the constraints that make test-time adaptation useful: existing models, no gradient updates, one strategy for every model and task. The strategy is \emph{rapidly adaptable}: a routing program is fitted to a task from the few demonstrations a prompt would contain anyway, by one forward pass per candidate and demonstration. And it is \emph{complementary to chain-of-thought}: a program re-allocates depth within one parallel forward pass whereas a chain spends sequential tokens, the two succeed on different queries, and a confidence-based arbitration composes them. We apply the strategy unchanged to seven post-trained models of four families and 7--31B parameters on MMLU~\citep{hendrycks2021mmlu} and a textual ARC-AGI~\citep{chollet2019arc}, trace the ability it exploits along the training trajectories of OLMo-3~\citep{olmo3} and SmolLM3~\citep{smollm3}, and analyse what helpful routes do to the residual stream, why others fail, and whether chain-of-thought does the same. Our main findings are:

\begin{enumerate}[label={},leftmargin=0pt,itemsep=0pt,topsep=2pt]
\item \finding{1} \textbf{Generalisable, training-free routing strategies already exist.} A program fitted on a task's demonstrations and arbitrated by the model's own confidence improves on the standard pass on every one of seven models and never hurts it, whereas trained routers reproduced on the same bed at best tie it (Table~\ref{tab:mmlu}). On ARC-AGI, programs fitted on a task's own two or three demonstrations lift a 7B model from $.27$ to $.43$ on an unseen split and a vote over them to $.54$ (Table~\ref{tab:arc}).
\item \finding{2} \textbf{Routing complements chain-of-thought.} The two succeed on different queries: composed by confidence, they exceed chain-of-thought alone on four of seven models and on average (Table~\ref{tab:mmlu}); on ARC-AGI the chain of the 7B model never terminates and degrades with its budget while routing doubles accuracy (Table~\ref{tab:arc}).
\item \finding{3} \textbf{The abilities routing exploits emerge early in pre-training, and most of the potential is still untapped.} An oracle over the standard answer, one fitted program and chain-of-thought reaches $.816$ macro accuracy on MMLU against $.692$ for chain-of-thought and $.660$ for the standard pass (Table~\ref{tab:mmlu}). Along the OLMo-3 trajectory, a vote over fitted programs adds $+.05$ to the standard pass at 25\% of pre-training and still $+.06$ at its end; the gain disappears at the first post-training stage, which saturates the model's confidence and entrenches the default computation, yet a reproducible headroom of $+.06$ to $+.24$ persists at every stage of OLMo-3 and SmolLM3 (Figure~\ref{fig:arc_cot}): post-training removes the ability to \emph{select} alternative computations, not the computations.
\item \finding{4} \textbf{Helpful routes are localised refinements that stay in distribution; failed routes leave it.} Helpful programs are overwhelmingly \emph{cycles}, one extra pass over a short block of early or middle layers, whose output stays close to what the full model computes; matched on problem and cost, unsuccessful programs drive the residual stream out of the distribution the following layers expect (Tables~\ref{tab:what} and~\ref{tab:ood}). At the decision points of sampled chains on the same problems, the trace of a helpful cycle is present but no more than that of a harmful one, and no more in chains that succeed than in chains that fail (Figure~\ref{fig:schema}): routing and reasoning are different levers on the same parameters.
\end{enumerate}

\section{Related Work}
\label{sec:related}

\paragraph{Test-time adaptation by generation.} Chain-of-thought~\citep{wei2022cot}, self-consistency~\citep{wang2023selfconsistency}, budget forcing~\citep{muennighoff2025s1} and test-time compute scaling~\citep{snell2025scaling} adapt a model to a problem by generating more tokens with a fixed computation; we study the complementary axis, in which the computation changes.

\paragraph{Learned recurrence and skipping.} \emph{Recurrent-depth} models share a block across depth and learn when to stop iterating it~\citep{dehghani2019universal,tan2023sut,giannou2023looped,yang2024looped,saunshi2025reasoning,geiping2025scaling,bae2025relaxed,bae2025mor,chen2025itt}; all train a new model whose depth policy is tied to its training distribution. \emph{Skip-only conditional computation} keeps ordinary parameters and learns which blocks to bypass: early exit and layer dropout~\citep{elbayad2020depthadaptive,fan2020layerdrop,elhoushi2024layerskip}, mixture-of-depths~\citep{raposo2024mod}, dynamic layer operations~\citep{tan2024dlo}, FlexiDepth~\citep{luo2025flexidepth}, and Router-Tuning/MindSkip~\citep{he2025mindskip}, whose routers skip \emph{attention sub-layers} of a frozen model, trading a fifth of the compute for a fraction of a point on likelihood-ranked benchmarks. Their aim is efficiency~\citep[see also][]{glavas2024dynamic,lawson2025middle}.

\paragraph{Joint skip/repeat programs over pretrained layers.} \citet{li2025cola} first skipped and looped the layers of pretrained models without training, showing by Monte-Carlo tree search (MCTS) with the answer in the loop that some execution order solves many problems the standard pass misses; Dr.LLM~\citep{heakl2025drllm} trains per-layer skip/execute/repeat routers on such labels, and \polar~\citep{li2026polar} formalises execution orders as \emph{programs}, discovers them by MCTS and trains a predictor over a separate embedding model to generate them. Both are supervised by task answers, and their transfer is narrow. Dr.LLM's routers, trained on ARC-Easy/Challenge and DART-Math with a router per backbone, \emph{lose} $0.85$ points on average across eight unseen benchmarks ($1.2$ outside mathematics) while saving computation; an ablation feeding the router the first-layer representation instead of the current state costs 4--9 points, so the rule depends on the evolving state rather than the task's surface. \polar's predictor gains on ASDiv, MAWPS and MMLU-Pro from mathematics-only supervision, but with one predictor per frozen backbone, an external encoder and no comparison with chain-of-thought: evidence for ``same model, new task'', not for a transferable routing principle. We keep \polar's program language, ask what a model-agnostic, training-free routing strategy reaches, and reproduce all three routers on our bed (\S\ref{sec:mmlu}). Finally, the \emph{superficial alignment hypothesis}~\citep{zhou2023lima,lin2024urial,yue2025limit} holds that post-training mainly teaches format over knowledge already in the base model; our trajectory experiment speaks to it from the side of internal computation.

\section{Methods}
\label{sec:methods}

Dynamic routing chooses, per input, which of a model's layers to execute and how often; we adopt the program-of-layers formalisation of \citet{li2026polar}, fit programs from demonstrations, and let confidence arbitrate between routing and reasoning.

\subsection{Programs of layers}
\label{sec:polar}

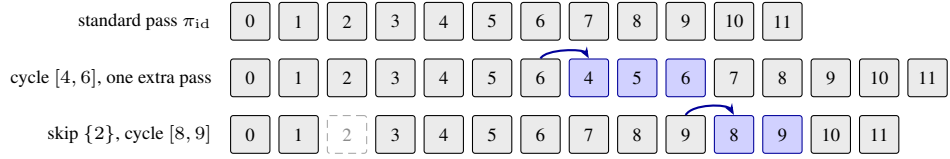
\begin{figure}[t]
\centering
\begin{tikzpicture}[
  layer/.style={draw, rounded corners=1pt, minimum width=5.2mm, minimum height=5mm, inner sep=0pt, font=\scriptsize},
  kept/.style={layer, fill=black!8},
  rep/.style={layer, fill=blue!18, draw=blue!60!black},
  skp/.style={layer, fill=white, draw=black!35, text=black!45, densely dashed},
  lbl/.style={font=\scriptsize, anchor=east},
  node distance=0mm and 1.2mm]
\node[lbl] (l0) at (0,0) {standard pass $\pi_{\mathrm{id}}$};
\foreach \i in {0,...,11} {\node[kept, right=of l0, xshift=\i*6.4mm] (a\i) {\i};}
\node[lbl] (l1) at (0,-0.75) {cycle $[4,6]$, one extra pass};
\foreach \i in {0,...,6} {\node[kept, right=of l1, xshift=\i*6.4mm] (b\i) {\i};}
\foreach \i/\j in {4/7,5/8,6/9} {\node[rep, right=of l1, xshift=\j*6.4mm] (b\j) {\i};}
\foreach \i/\j in {7/10,8/11,9/12,10/13,11/14} {\node[kept, right=of l1, xshift=\j*6.4mm] (b\j) {\i};}
\draw[-{Stealth[length=1.5mm]}, blue!60!black, thick] (b6.north) to[out=60,in=120,looseness=0.6] (b7.north);
\node[lbl] (l2) at (0,-1.5) {skip $\{2\}$, cycle $[8,9]$};
\foreach \i/\j in {0/0,1/1} {\node[kept, right=of l2, xshift=\j*6.4mm] (c\j) {\i};}
\node[skp, right=of l2, xshift=2*6.4mm] (c2) {2};
\foreach \i/\j in {3/3,4/4,5/5,6/6,7/7,8/8,9/9} {\node[kept, right=of l2, xshift=\j*6.4mm] (c\j) {\i};}
\foreach \i/\j in {8/10,9/11} {\node[rep, right=of l2, xshift=\j*6.4mm] (c\j) {\i};}
\foreach \i/\j in {10/12,11/13} {\node[kept, right=of l2, xshift=\j*6.4mm] (c\j) {\i};}
\draw[-{Stealth[length=1.5mm]}, blue!60!black, thick] (c9.north) to[out=60,in=120,looseness=0.7] (c10.north);
\end{tikzpicture}
\vspace{-2mm}
\caption{Programs of layers for a 12-layer model. Boxes are layer applications in execution order; blue boxes re-apply a contiguous block (a \emph{cycle}), the dashed box is a skipped layer. Every program runs the unchanged pretrained weights and is read out in one forward pass.}
\label{fig:program}
\vspace{-3mm}
\end{figure}

Let a decoder-only transformer have layers $f_0,\dots,f_{D-1}$ and residual stream $h$. Standard inference executes the identity program $\pi_{\mathrm{id}} = (0,1,\dots,D-1)$, i.e.\ $h \leftarrow h + f_\ell(h)$ for $\ell$ in that order. A \emph{program of layers}~\citep{li2026polar} is any finite sequence $\pi = (\ell_1,\dots,\ell_T)$ over $\{0,\dots,D-1\}$ executed in the same way; a layer absent from $\pi$ is \emph{skipped}, a layer occurring more than once is \emph{repeated}. Following \citet{li2025cola}, we restrict repeats to \emph{cycles}: a contiguous block $[b_0, b_1]$ of at most $k_{\max}$ layers executed $r$ extra times, possibly combined with skips (Figure~\ref{fig:program}). Repeated layers need a key--value cache indexed by \emph{program position}; we implement this per family and verify that the identity program is bit-exact with the vanilla model (Appendix~\ref{app:exec}). A program costs at most a few percent more layer applications than a standard pass and is read out from a \emph{single} pass: the answer scaffold (``\texttt{The final answer is \textbackslash boxed\{}'') is pre-filled so that the model commits to an answer immediately, without a chain of thought.

\subsection{Adaptable program discovery from demonstrations}
\label{sec:fewshot}

\citet{li2026polar} discover programs per problem with the ground-truth answer in the loop and train a predictor on thousands of such problems. For a new task we assume instead only the labelled demonstrations that a few-shot prompt would contain anyway, and no other supervision; like every few-shot method, this presupposes that a test query is posed with demonstrations of its \emph{own} task (\S\ref{sec:discussion}). Each demonstration is posed in turn as the query with the remaining demonstrations in context, and a candidate program is scored by the probability it assigns to the gold answer, teacher-forced after the answer scaffold in one pre-fill forward pass with no decoding:
\needspace{9\baselineskip}
\begin{itemize}
  \item $\mathcal{D}=\{(x_i,y_i)\}_{i=1}^{n}$: the task's $n$ labelled demonstrations; $n{=}10$ for MMLU, $n{=}3$ for ARC-AGI
  \item $x_i$: the prompt of demonstration $i$; $y_i$: its gold answer.
  \item $p_\pi(y_i \mid \mathcal{D}_{-i}, x_i)$: the probability that the program $\pi$ assigns to gold label $y_i$
  \item Ranking $R$ of the program $\pi$ is defined as:
\begin{equation}
\label{eq:evidence}
  R(\pi) \;=\; \frac{1}{n}\sum_{i=1}^{n} p_\pi\!\left(y_i \,\middle|\, \mathcal{D}_{-i},\, x_i\right)
\end{equation}
\end{itemize}
On multiple-choice tasks $p_\pi$ is the probability of the gold option normalised over the options; on free-form tasks the length-normalised probability of the gold text. The score must be continuous and contrastive (Appendix~\ref{app:negative}).

\paragraph{Search, single program, vote.} Candidates are proposed by the tree search of \citet{li2026polar} (block actions \emph{keep}, \emph{skip}, \emph{repeat}; 300 rollouts per task; blocks of at most four layers, at most two extra passes; Appendix~\ref{app:setup}) with $R(\pi)$ as reward, run on the model under test, and the ten best are kept. \pp\ executes the best candidate; \vote$_{15}$ fits 15 members on overlapping 8-of-10 subsets of the demonstrations, each with its own search, and takes the weighted majority of their answers, each member weighted by its fitness on its two held-out demonstrations and, in the \emph{soft} vote, by its confidence in its answer (Eq.~\ref{eq:conf}).

\subsection{Confidence-based selection}
\label{sec:conf}

A fitted program is not right for every query, nor is chain-of-thought. We treat standard inference (\dir), the fitted program (\pp) and chain-of-thought (\cotarm) as \emph{arms} that each produce an answer $a$ with the model's confidence in it,
\begin{equation}
\label{eq:conf}
  c(a) \;=\; \Big(\textstyle\prod_{t=1}^{|a|} p(a_t \mid a_{<t}, \cdot)\Big)^{1/|a|},
\end{equation}
the geometric mean of the token probabilities of the emitted answer (for \cotarm, read at the answer position after the chain). \cs\ answers with the arm of highest confidence, ties broken in the fixed order \cotarm\ $\succ$ \dir\ $\succ$ \pp. Its one-parameter relative, \mrg, keeps a \emph{default} arm and lets a challenger take the query only if its confidence exceeds the default's by a margin $\tau$, chosen by 5-fold cross-fitting over tasks, so that every reported number is out of fold. We use \mrg\ with \cotarm\ as default (the composition of programs with reasoning) and with \dir\ as default and \pp\ as the only challenger (routing without reasoning); further variants are in Appendix~\ref{app:variants}.

\subsection{Oracles: hindsight and transferable}
\label{sec:oracle}

To separate what the model \emph{can} compute from what a rule \emph{does} select, we report an \orc\ that, per query, is correct if any of the arms is; the gap between \orc\ and the best rule is the price of selection. A hindsight oracle over \emph{many} programs is a multiple-comparisons trap: on ARC-AGI two thirds of the programs the search called valid did not survive held-out option orderings. We therefore report oracles over three arms only, over held-out orderings, or next to their independence null, and complement them with a \torc: every arm is sampled a second, independent time and a program counts for a query only if it is correct in \emph{both} draws, which removes the sampling luck a hindsight oracle harvests (a conservative floor, since a good but stochastic program is penalised twice) and is the most a selection rule can reach.

\section{Experiments}
\label{sec:experiments}

\paragraph{Setup.} Seven post-trained models covering four families, dense and mixture-of-experts architectures and 28--60 layers: DeepSeek-R1-Distill-Qwen-7B~\citep{deepseek2025r1}, Llama-3.1-8B-Instruct~\citep{llama3}, Ministral-3-8B-Reasoning~\citep{ministral3}, Qwen3-8B~\citep{qwen3}, Qwen3.5-9B~\citep{qwen35}, Gemma-4-26B-A4B and Gemma-4-31B~\citep{gemma4}, each with its own chat template and card sampling (Appendix~\ref{app:setup}). \emph{MMLU bed:} 49 subjects $\times$ 20 test questions (980 queries) and, per subject, 10 further questions that serve as demonstrations: the in-context examples of every test query and, in leave-one-out form, the 490 \emph{fit} items of Eq.~\ref{eq:evidence}. The model answers with the \emph{text} of the option, scored by normalised exact match, so a routed pass has to produce the answer; \dir\ and \pp\ generate at most 120 tokens after the scaffold, \cotarm\ reasons with a 10k-token budget; we report macro accuracy over subjects. \emph{Baselines:} MindSkip~\citep{he2025mindskip}, Dr.LLM~\citep{heakl2025drllm} and the \polar\ predictor~\citep{li2026polar}, each in the \emph{transfer} configuration of its paper (routers trained on the paper's data, applied to MMLU) and in a \emph{few-shot} configuration that trains the same routers on our 490 fit items, so that they see exactly what \pp\ sees (Appendix~\ref{app:baselines}).

\subsection{MMLU: one strategy, seven models}
\label{sec:mmlu}

\begin{table}[t]
\caption{Accuracy on MMLU (49 tasks $\times$ 20 queries, macro accuracy). \dir: standard pass; \cotarm: 10k-token chain-of-thought; \cs: highest-confidence arm; \orc: per-query best of the three arms. Bold/underline: \textbf{best}/\underline{second-best} method. 
}
\label{tab:mmlu}
\begin{center}
\scriptsize
\setlength{\tabcolsep}{3pt}
\begin{tabular}{lcccccccc}
\toprule
 & DeepSeek-R1 & Llama-3.1 & Ministral-3 & Qwen3 & Qwen3.5 & Gemma-4 & Gemma-4 & \\
Method & 7B & 8B & 8B & 8B & 9B & 26B-A4B & 31B & avg \\
\midrule
\dir\ (standard pass)             & .242 & .497 & .618 & .747 & .795 & .842 & .882 & .660 \\
\cotarm\ (10k tokens)             & .294 & \best{.629} & .554 & .724 & .823 & \best{.895} & \best{.929} & .692 \\
\midrule
MindSkip \citep{he2025mindskip}      & .082 & .350 & .405 & .156 & .618 & .765 & .740 & .445 \\
Dr.LLM \citep{heakl2025drllm}        & .249 & .349 & .621 & .687 & .786 & .789 & .882 & .623 \\
\polar\ \citep{li2026polar} & .231 & .501 & .622 & .747 & .787 & .850 & .882 & .660 \\
\midrule
\cs\ \{\cotarm,\pp\}           & \best{.329} & .524 & \best{.628} & \best{.750} & \second{.837} & .886 & .915 & \second{.696} \\
\ \ + conf. margin  & \second{.327} & \second{.625} & \second{.627} & \best{.750} & \best{.841} & \second{.893} & \second{.922} & \best{.712} \\
\midrule
\textit{\orc}   & \textit{.530} & \textit{.774} & \textit{.783} & \textit{.855} & \textit{.901} & \textit{.928} & \textit{.943} & \textit{.816} \\
\bottomrule
\end{tabular}
\end{center}
\vspace{-5mm}
\end{table}

\paragraph{Finding 1a: arbitrated by confidence, a fitted program never loses to the standard pass.} Table~\ref{tab:mmlu} applies the strategy of \S\ref{sec:methods} unchanged to the seven models; Table~\ref{tab:variants} (Appendix~\ref{app:variants}) lists every single arm and selection rule. A single program per subject is a coarse instrument: \pp\ alone beats \dir\ on two models and loses on five, by up to 13 points on Qwen3-8B. The search tells why: on the Gemmas and Ministral it found no program beating the standard pass on more than three of 49 subjects, whereas on DeepSeek and Llama it found rich structure on most subjects, which transferred to the test queries for DeepSeek and not for Llama (demonstration overfit, the failure mode of every fitted router below). Arbitration makes this safe by construction: \mrg\ with \dir\ as default and \pp\ as the only challenger is at or above \dir\ on every model ($+1.4$ on average, worst cell $-0.1$) and captures the whole repair where it exists (DeepSeek $+6.7$, McNemar $p{<}10^{-7}$).

\paragraph{Finding 2: routing is complementary to chain-of-thought, and their composition is superior to it.} Routing and reasoning succeed on different queries: \cs\ over the three arms beats \emph{every} single arm on four of seven models, including the one on which \cotarm\ is six points \emph{below} the direct pass (Ministral) and the one on which \cotarm\ is the best arm (Qwen3.5). Where \cotarm\ dominates by a wide margin (Llama, both Gemmas) max-confidence loses to it, because the Gemmas' confidences saturate near $.997$ and Llama's arms have mismatched confidence scales (\S\ref{sec:ladder}). The margin rule with \cotarm\ as default (``$+$ conf.\ margin'' in Table~\ref{tab:mmlu}) repairs exactly this: it is the best method on average ($+1.9$ over always-\cotarm, $+5.2$ over \dir), positive on four of seven models and never more than $0.6$ below \cotarm; per domain it pays on recall-flavoured subjects and costs on a few calculation-heavy ones (Appendix~\ref{app:variants}).

\paragraph{Finding 1b: aggregate programs for weak models, arbitrate regimes for strong ones.} On the weakest model the right strategy is not to arbitrate but to \emph{aggregate}: the soft vote over 15 programs fitted on different demonstration subsets lifts DeepSeek from $.242$ to $.412$, 12 points above its \cotarm, and the same members applied \emph{under} \cotarm\ compose with it (Table~\ref{tab:ablation}). The gain is bagging diversity, not voting, and it inverts with capability (on Qwen3-8B the vote lands below \dir).

\paragraph{Untapped potential: the oracle shows what selection leaves behind.} One of the three arms is correct on $.816$ of the queries on average, 12 points above \cotarm\ and 16 above \dir, on every model; a third of that headroom is not even task-level (an oracle choosing one arm per subject reaches $.742$, Table~\ref{tab:variants}). What limits us is not the existence of a better computation but recognising it: confidence can \emph{veto} a switch, not navigate to the best arm.

\paragraph{Where trained routers stand.} Table~\ref{tab:mmlu} shows MindSkip in its paper's transfer configuration and Dr.LLM and the \polar\ predictor trained on our 490 fit items (both configurations for all three, with caveats on the Gemma arms, in Table~\ref{tab:baselines}). In their own transfer configurations none of the baselines improves on the standard pass: MindSkip's routers cost 8--59 points on every model (its published robustness rests on likelihood ranking of the options, not on \emph{producing} the answer), and the \polar\ predictor trained on DART-Math pools lands below \dir. Given the same 490 demonstrations \pp\ uses, the \polar\ predictor converges to ``do not route'' and ties \dir, and Dr.LLM's routers learn the search's structure and pay for it at test time (Llama $-14.8$, Qwen3 $-6.0$): matched supervision buys trained routers safety, never gains, because none has an arbitration layer. \emph{Cost:} \pp\ adds one pre-fill per candidate and demonstration at fitting time and a few extra layer applications at test time, where a 10k-token chain takes thousands of sequential steps.

\subsection{ARC-AGI: rapid adaptation from a task's own demonstrations}
\label{sec:arc}

The severest test of an inference-time method is a task on which the model with chain-of-thought fails. We use a textual, multiple-choice rendering of ARC-AGI-1~\citep{chollet2019arc} (grids as nested lists following \citet{arcvl2025}; four candidate output grids); a program is correct on a task only if it picks the gold grid under all three stored held-out option orderings ($k{=}3$ consistency: uniform guessing scores $1.6\%$, the strongest content-based guessing rule $22\%$). The setup is the few-shot one of \S\ref{sec:fewshot} with each task's \emph{own} two or three demonstrations, posed one at a time as queries with the others in context (Eq.~\ref{eq:evidence}): programs are proposed by the model-specific tree search, \pp\ takes the best-fitting program, \vote$_{15}$ votes over programs fitted on demonstration subsets, and nothing about the evaluation split or the test query is used (Appendix~\ref{app:arc}).

\begin{table}[t]
\caption{Textual ARC-AGI-1, DeepSeek-R1-Distill-Qwen-7B, unseen evaluation split (177 tasks), $k{=}3$ demonstrations. Per-task adaptation uses only the task's own demonstrations; fixed programs are selected on the training split.}
\label{tab:arc}
\begin{center}
\scriptsize
\setlength{\tabcolsep}{3.5pt}
\begin{tabular}{lcl}
\toprule
Strategy on the unseen split & accuracy & vs.\ standard pass \\
\midrule
random guessing                                              & .200 & -- \\
standard pass                                              & .271 & -- \\
\cotarm, best-performing thinking budget                   & .350 & $+.03$ on the same 60 tasks; longer budgets fall to $.300$ and $.200$ \\
\pp\ fitted on the task's own demonstrations (Eq.~\ref{eq:evidence}) & .429 & better on 34 tasks, worse on 6 ($p{=}8{\times}10^{-6}$) \\
\ \ + Voting $R_\pi$ over programs from demonstrations  & \textbf{.540} & better on 22, worse on 3 vs.\ \pp \\
\midrule
\textit{Oracle: best test score among the found programs}           & \textit{.605} & \\
\bottomrule
\end{tabular}
\end{center}
\vspace{-5mm}
\end{table}

\paragraph{Finding 1c: two or three demonstrations suffice to double accuracy.} Per-task adaptation lifts DeepSeek-R1-Distill-Qwen-7B from $.271$ to $.429$ with no validation set and no training (Table~\ref{tab:arc}): the demonstration signal chooses 23 distinct programs across tasks and saturates after about six demonstration measurements, so the remaining gap to the existence bound is a validity ceiling of that signal rather than estimation noise (Appendix~\ref{app:arc}); the 15-member vote over programs fitted on demonstration subsets pushes through it to $.540$, twice the standard pass, for a handful of pre-fill passes per task. Reasoning does not help here (Table~\ref{tab:arc}): the best thinking budget, 256 tokens, gains three points over the standard pass on the 60 training tasks of the budget sweep, longer budgets fall below it, and the chain never terminates at any budget, so what degrades is the forced answer's agreement with itself across reshuffled options (Appendix~\ref{app:arc}); the same pathology appears on BIG-bench's program-induction tasks (Appendix~\ref{app:bb}).

\paragraph{A window, not a low-accuracy regime.} The same recipe repairs nothing for Qwen3-8B on the same tasks ($.733$ standard pass, $.699$ fitted program and vote; Appendix~\ref{app:arc}), as on the strong post-trained models of \S\ref{sec:mmlu}. Repair needs a model competent enough for demonstration evidence to carry signal and weak enough for the default order to misexecute, a property of the \emph{model} that \S\ref{sec:ladder} locates on the training axis.

\subsection{Along the training trajectory: when do the abilities routing exploits emerge?}
\label{sec:ladder}

\begin{figure}[t]
\centering
\begin{minipage}[t]{0.49\linewidth}
\centering
\begin{tikzpicture}
\begin{axis}[
  width=\linewidth, height=3.1cm, title={\scriptsize OLMo-3-7B}, title style={yshift=-1.5mm},
  xlabel={training stage}, ylabel={MMLU macro accuracy},
  symbolic x coords={25\%,50\%,100\%,base,SFT,DPO,RLVR}, xtick=data,
  x tick label style={font=\tiny},
  ymin=0.15, ymax=0.95, grid=major, grid style={black!10},
  legend to name=legladder, legend columns=5, legend style={draw=none, fill=none, font=\tiny, cells={anchor=west}, /tikz/every even column/.append style={column sep=6pt}},
  every axis plot/.append style={thick, mark size=1.5pt},
  tick label style={font=\scriptsize}, label style={font=\scriptsize}]
\addplot+[color=black!40, dashed, mark=o] coordinates {(25\%,.803) (50\%,.833) (100\%,.855) (base,.866) (SFT,.842) (DPO,.831) (RLVR,.827)};
\addlegendentry{hindsight \orc}
\addplot+[color=red!70!black, mark=diamond*] coordinates {(25\%,.600) (50\%,.672) (100\%,.665) (base,.768) (SFT,.791) (DPO,.794) (RLVR,.784)};
\addlegendentry{\torc}
\addplot+[color=blue!70!black, mark=square*] coordinates {(25\%,.448) (50\%,.484) (100\%,.531) (base,.592) (SFT,.599) (DPO,.612) (RLVR,.595)};
\addlegendentry{soft-\vote$_{15}$}
\addplot+[color=black, mark=*] coordinates {(25\%,.392) (50\%,.470) (100\%,.487) (base,.533) (SFT,.602) (DPO,.600) (RLVR,.601)};
\addlegendentry{\dir}
\addplot+[color=blue!40, mark=triangle*] coordinates {(25\%,.376) (50\%,.412) (100\%,.457) (base,.518) (SFT,.564) (DPO,.564) (RLVR,.524)};
\addlegendentry{\pp}
\end{axis}
\end{tikzpicture}
\end{minipage}\hfill
\begin{minipage}[t]{0.49\linewidth}
\centering
\begin{tikzpicture}
\begin{axis}[
  width=\linewidth, height=3.1cm, title={\scriptsize SmolLM3-3B}, title style={yshift=-1.5mm},
  xlabel={training stage}, ylabel={MMLU macro accuracy},
  symbolic x coords={5\%,15\%,25\%,50\%,st.3,base,SFT,APO,final}, xtick=data,
  x tick label style={font=\tiny},
  ymin=0.15, ymax=0.95, grid=major, grid style={black!10},
  every axis plot/.append style={thick, mark size=1.5pt},
  tick label style={font=\scriptsize}, label style={font=\scriptsize}]
\addplot+[color=black!40, dashed, mark=o] coordinates {(5\%,.625) (15\%,.675) (25\%,.743) (50\%,.740) (st.3,.660) (base,.582) (SFT,.652) (APO,.740) (final,.765)};
\addplot+[color=red!70!black, mark=diamond*] coordinates {(5\%,.370) (15\%,.439) (25\%,.518) (50\%,.525) (st.3,.456) (base,.366) (SFT,.548) (APO,.653) (final,.661)};
\addplot+[color=black, mark=*] coordinates {(5\%,.252) (15\%,.262) (25\%,.332) (50\%,.362) (st.3,.263) (base,.204) (SFT,.484) (APO,.556) (final,.549)};
\addplot+[color=blue!40, mark=triangle*] coordinates {(5\%,.224) (15\%,.242) (25\%,.312) (50\%,.310) (st.3,.266) (base,.236) (SFT,.479) (APO,.523) (final,.551)};
\end{axis}
\end{tikzpicture}
\end{minipage}\\[-1mm]
\pgfplotslegendfromname{legladder}
\vspace{-2mm}
\caption{Finding 3: two training trajectories on the MMLU bed. Standard pass, single fitted program, on OLMo-3 the vote over 15 fitted programs, the hindsight per-query oracle over the standard pass and ten searched programs (first sampling draw), and the \torc\ (program correct in both of two independent draws). The vote's gain on OLMo-3 vanishes at SFT; the reproducible headroom persists at every stage, while the hindsight oracle's inflation by sampling luck shrinks with post-training. SmolLM3's stage-3 and base checkpoints run under the raw scaffold, which depresses their standard pass (answer-format compliance, not capability).}
\label{fig:arc_cot}
\vspace{-4mm}
\end{figure}
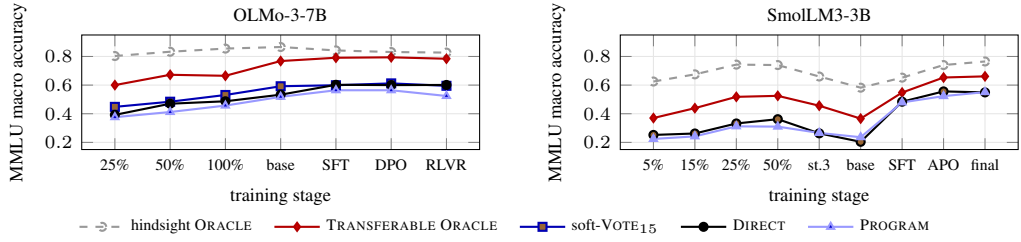

OLMo-3~\citep{olmo3} and SmolLM3~\citep{smollm3} release intermediate checkpoints of their pre-training runs together with their post-training stages, which lets us run the identical protocol along two trajectories: OLMo-3-7B at $\approx$5, 15, 25, 50 and 100\% of stage-1 pre-training, the base model, and the supervised fine-tuning (SFT), direct preference optimisation (DPO) and reinforcement learning with verifiable rewards (RLVR) checkpoints of its instruct model; SmolLM3-3B at matched fractions of stage 1, the stage-3 and base checkpoints, and its SFT, preference-optimised (APO) and final instruct checkpoints (no \cotarm\ arm, since base checkpoints cannot reason). Per rung we fit programs per subject as in \S\ref{sec:fewshot} and evaluate \dir, \pp, the two-arm \cs\ and \mrg, on OLMo-3 the 15-member bagged vote, and the hindsight oracle over \dir\ and the ten searched candidates next to its \torc\ from an independent second sampling draw (\S\ref{sec:oracle}). Figure~\ref{fig:arc_cot} shows the result; Tables~\ref{tab:ladder} and~\ref{tab:replication} (Appendix~\ref{app:ladder}) have every rung and rule.

\paragraph{Finding 3a: early emergence, plateau, and a boundary at the first post-training stage.} On OLMo-3 the gain of the bagged vote over the standard pass is already present at the earliest rung on which we ran it, $+5.7$ points at 25\% of pre-training, where the standard pass answers 39\% of the queries: routability does not wait for the model to be good. From there to the base model the standard pass improves by 14 points while the vote's gain stays essentially constant: the ability routing adds is not what pre-training is adding. At SFT the standard pass gains 7 points and the vote's advantage drops to zero, where it stays through DPO and RLVR. On SmolLM3 (no vote run) the single program tracks the standard pass at every rung.

\paragraph{Finding 3b: post-training saturates confidence and entrenches the default computation.} Two readouts change abruptly at the same stage (Table~\ref{tab:ladder}). The median answer confidence of the standard pass is flat near $.6$ through pre-training while accuracy climbs 10 points, then jumps to $.75$ at the base model and to $.97/.98/.99$ across SFT/DPO/RLVR (SmolLM3: $.64$ at the base, $.91$ at the instruct checkpoint): \emph{confidence saturation is manufactured by post-training}, which is what made the Gemmas' arms tie in \S\ref{sec:mmlu}. And the fitted program's confidence relative to the standard pass flips from even during pre-training to favouring the default computation after any alignment stage. This is the internal-computation counterpart of the superficial alignment hypothesis~\citep{zhou2023lima,lin2024urial}: the ability that post-training surfaces in the default pass is present, and reachable by re-ordering the same layers, long before.

\paragraph{Finding 3c: the headroom is real and reproducible, but the hindsight oracle overstates it.} The hindsight oracle over eleven arms is large at every rung of both ladders ($.58$--$.87$) but sits \emph{below} the independence null of eleven answerers with the same marginal accuracies (Tables~\ref{tab:ladder} and~\ref{tab:replication}), and a second independent sampling draw shows how much of it is luck: $+.16$ to $+.26$ points on pre-training rungs, $+.04$ to $+.10$ after post-training, as sampling variance collapses with confidence. What survives both draws, the \torc, stays $+.18$ to $+.24$ above the standard pass on OLMo-3 and $+.06$ to $+.19$ on SmolLM3 at every stage. Post-training does not remove the alternative computations; it removes our ability to \emph{select} them (Finding 3b).

\begin{table}[t]
\caption{Finding 4. \textbf{(a)} What helpful programs are and do (DeepSeek-R1-Distill-Qwen-7B; GSM8K search, matched fix/break pairs and fixed-cycle scan; verified vs.\ search-rejected ARC-AGI-1 programs). \textbf{(b)} The trace of a helpful cycle at the decision points of chain-of-thought (MMLU, Qwen3.5-9B, 213 split prefixes over 32 items; GSM1k, DeepSeek-R1-Distill-Qwen-7B, 74 problems); brackets: 95\% bootstrap intervals over items; offset-0 excess: cosine between $C$ and the chain's update at the cycle's own depth minus the permuted-item null (raw / shared components removed). AUC: area under the ROC curve; KL: Kullback--Leibler divergence to the full model's output; full statistics in Appendix~\ref{app:cotloops}.}
\label{tab:what}
\begin{center}
\scriptsize
\setlength{\tabcolsep}{2.5pt}
\begin{tabular}{lll}
\toprule
\multicolumn{3}{l}{\textbf{(a) Finding 4a: what helpful programs are and do}} \\
Question & Statistic & Value \\
\midrule
Repairing program exists?         & standard-pass misses corrected by some program (GSM8K) & 61\% (oracle .70 vs.\ .245) \\
Are repairs cycles?               & repairing programs containing a repeat (GSM8K / ARC)            & 99\% / 56 of 56 (pure skips 6/634, 0) \\
Where do valid programs cycle?    & layers 1--2 / 11--18 / last two, valid vs.\ rejected (ARC)       & .82/.31/.00 vs.\ .62/.15/.14 \\
How many extra passes?            & fixed cycle with 1 / 2 / 3 / 4 extra passes (GSM8K)             & .258/.192/.115/.070 (standard .300) \\
Does the refinement stay close?   & validity scorer over 31 label-free features                     & AUC .80; top: KL proximity \\
Is the fix direction generic?     & cosine with the block's own first-pass update                   & $+.65$ (other block $+.09$) \\
Is repair decodable?              & per-problem probe / across problems / across depth              & AUC .76 / .55 / .54 (permuted .51) \\
\bottomrule
\end{tabular}

\vspace{1.5mm}
\begin{tabular}{lll}
\toprule
\multicolumn{3}{l}{\textbf{(b) Finding 4b: the trace of a helpful cycle at the decision points of chain-of-thought}} \\
Question (statistic) & MMLU / Qwen3.5-9B & GSM1k / DeepSeek-R1 7B \\
\midrule
Is the trace in the chain at the matching depth? (offset-0 excess) & $+.17$ raw / $+.25$ residualised & $+.17$ / $+.18$ \\
Effective dimensions of the perturbation (1{,}902 fixing cycles) & $\approx 9.5$ (top 1: 29\%, top 5: 55\%) & -- \\
Do different items use the same coordinates? (top-64 overlap)  & 20.8 of 64 ($20\times$ chance) & -- \\
Stronger for helpful than for harmful cycles? (paired within item) & $-.04$ $[-.09, +.01]$ & $+.004$ $[-.07, +.08]$ \\
Stronger in chains that succeed? (within item) & $+.003$ $[-.008, +.014]$ & $-.001$ $[-.005, +.003]$ \\
Do the fix--break and correct--wrong axes align? (branching token) & $\le .0025$ cosine over 1.3M pairs & -- \\
Same cycle on the reasoning prompt: same vector? & cosine $+.07$; label survives (AUC .75) & -- \\
\bottomrule
\end{tabular}
\end{center}
\vspace{-5mm}
\end{table}

\section{Analyses}
\label{sec:analyses}

\subsection{What do the helpful programs do?}
\label{sec:what}

\begin{table}[t]
\caption{Do unsuccessful programs leave the distribution the following layers expect? DeepSeek-R1-Distill-Qwen-7B on DART-Math: per problem, one \emph{correct} and one \emph{wrong} program that both contain a cycle and are matched on total and repeated layer passes (419 pairs), so loop presence and compute are held fixed. Each program's final state (the input to the final norm and head) is scored against the full-depth reference cloud in a 256-dimensional principal subspace. Brackets: 95\% bootstrap intervals; AUC: area under the ROC curve for wrong vs.\ correct. Controls: the same scoring on unmatched pairs.}
\label{tab:ood}
\begin{center}
\scriptsize
\setlength{\tabcolsep}{4pt}
\begin{tabular}{lcccrc}
\toprule
Distance to the reference cloud & wrong & correct & paired $\Delta$ (wrong $-$ correct) & wrong further & AUC \\
\midrule
cosine (direction; survives the pre-norm) & .261 & .187 & $+.069$ $[+.050, +.081]$ & 69\% & .654 \\
Mahalanobis (magnitude; divided out by the norm) & 13.92 & 13.37 & $+.46$ $[-.07, +1.13]$ & 54\% & .554 \\
\midrule
control: cycle vs.\ no-cycle pairs & \multicolumn{5}{l}{gap follows loop presence, not correctness (fix 0\%, break 100\% of pairs)} \\
control: cycle-on states, not count-matched & \multicolumn{5}{l}{AUC .52} \\
\bottomrule
\end{tabular}
\end{center}
\vspace{-5mm}
\end{table}

\paragraph{Finding 4a: helpful routes are single extra passes that stay in distribution; failed routes leave it.} Table~\ref{tab:what}(a) collects the evidence across beds. The programs that repair the standard pass are cycles: on GSM8K, where a direct-answer search corrects most of the standard pass's misses, practically every repairing program contains a repeat, and on ARC-AGI none of the verified repairs is a pure skip. Against search-rejected programs on the same tasks, valid ARC programs cycle the first two layers and the middle band more often, \emph{never} cycle the last two layers, and use two passes rather than three; the motif is a property of the model rather than of the task. The scan of fixed cycles settles how many repeats: one extra pass preserves the standard pass's accuracy, further passes destroy it, because the looped block does not contract to a fixed point but \emph{expands} the state along one dominant direction. What separates valid from invalid programs is not any block-local geometry but the proximity, in the readout, of the program's one-step output distribution to the full model's, and Table~\ref{tab:ood} makes the geometry of this precise: for the same problem and matched layer-pass counts, the wrong program's output lies further, in \emph{direction}, from the cloud of states the final stage normally receives, while its magnitude does not, which is what a pre-norm network divides out before the next layer sees it; pairs that are not matched show only a loop-presence artifact. Helpful programs refine the state within the distribution the following layers can process; harmful ones leave it. Otherwise whether a cycle helps is decodable \emph{within} a problem but transfers neither across problems nor across depth, because the extra pass mostly re-applies what the block wrote the first time: repair is a problem-specific property of a generic operation.

\subsection{Does chain-of-thought perform the same refinement?}
\label{sec:cotloops}

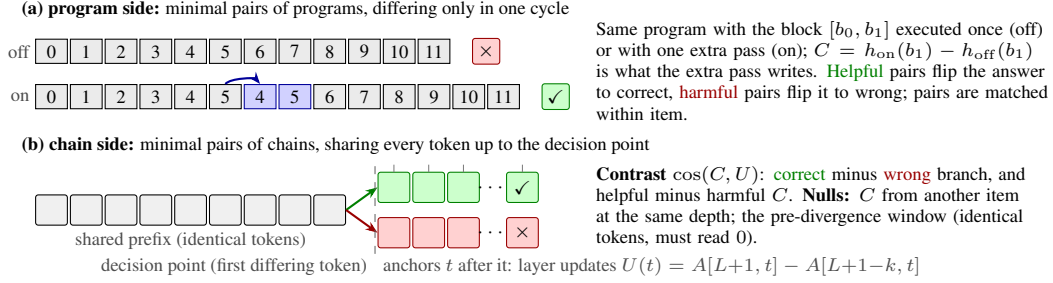
\begin{figure}[t]
\centering
\begin{tikzpicture}[
  font=\scriptsize,
  tok/.style={draw, rounded corners=1pt, minimum width=4.2mm, minimum height=4mm, inner sep=0pt, fill=black!6},
  ok/.style={tok, fill=green!20, draw=green!50!black},
  bad/.style={tok, fill=red!18, draw=red!60!black},
  lay/.style={draw, minimum width=4.2mm, minimum height=3.2mm, inner sep=0pt, fill=black!8},
  rep/.style={lay, fill=blue!18, draw=blue!60!black},
  arr/.style={-{Stealth[length=1.6mm]}, thick},
  node distance=0mm and 1mm]
\node[anchor=west, inner sep=0pt] at (0,2.35) {\textbf{(a) program side:} minimal pairs of programs, differing only in one cycle};
\node[anchor=east, inner sep=1pt, black!70] at (0.15,1.8) {off};
\foreach \i in {0,...,11} {\node[lay] (O\i) at (0.4+\i*0.46,1.8) {\i};}
\node[bad, minimum width=3.6mm, minimum height=3.4mm] at (6.15,1.8) {$\times$};
\node[anchor=east, inner sep=1pt, black!70] at (0.15,1.2) {on};
\foreach \i in {0,...,5} {\node[lay] (N\i) at (0.4+\i*0.46,1.2) {\i};}
\node[rep] (R4) at (0.4+6*0.46,1.2) {4}; \node[rep] (R5) at (0.4+7*0.46,1.2) {5};
\foreach \i/\j in {6/8,7/9,8/10,9/11,10/12,11/13} {\node[lay] (N\j) at (0.4+\j*0.46,1.2) {\i};}
\node[ok, minimum width=3.6mm, minimum height=3.4mm] at (7.07,1.2) {\checkmark};
\draw[arr, blue!60!black] (N5.north) to[out=50,in=130,looseness=0.6] (R4.north);
\node[anchor=west, text width=6.0cm, inner sep=0pt] at (7.6,1.55) {Same program with the block $[b_0,b_1]$ executed once (off) or with one extra pass (on); $C = h_{\mathrm{on}}(b_1) - h_{\mathrm{off}}(b_1)$ is what the extra pass writes. \textcolor{green!50!black}{Helpful} pairs flip the answer to correct, \textcolor{red!60!black}{harmful} pairs flip it to wrong; pairs are matched within item.};
\node[anchor=west, inner sep=0pt] at (0,0.55) {\textbf{(b) chain side:} minimal pairs of chains, sharing every token up to the decision point};
\foreach \i in {0,...,8} {\node[tok] (P\i) at (0.4+\i*0.46,-0.3) {};}
\node[black!70, inner sep=0pt] at (2.24,-0.7) {shared prefix (identical tokens)};
\node[ok] (A1) at (4.93,0.0) {}; \node[ok] (A2) at (5.39,0.0) {}; \node[ok] (A3) at (5.85,0.0) {}; \node[inner sep=0pt] at (6.23,0.0) {$\cdots$}; \node[ok] (A5) at (6.63,0.0) {\checkmark};
\node[bad] (B1) at (4.93,-0.6) {}; \node[bad] (B2) at (5.39,-0.6) {}; \node[bad] (B3) at (5.85,-0.6) {}; \node[inner sep=0pt] at (6.23,-0.6) {$\cdots$}; \node[bad] (B5) at (6.63,-0.6) {$\times$};
\draw[arr, green!50!black] (P8.east) -- (A1.west);
\draw[arr, red!60!black] (P8.east) -- (B1.west);
\draw[dashed, black!60] (4.68,0.28) -- (4.68,-0.9);
\node[black!70, anchor=east, inner sep=0pt] at (4.58,-1.05) {decision point (first differing token)};
\foreach \x in {4.93,5.39,5.85,6.63} {\draw[black!50] (\x,0.2) -- (\x,0.3);}
\node[black!70, anchor=west, inner sep=0pt] at (4.78,-1.05) {anchors $t$ after it: layer updates $U(t)=A[L{+}1,t]-A[L{+}1{-}k,t]$};
\node[anchor=west, text width=6.0cm, inner sep=0pt] at (7.6,-0.25) {\textbf{Contrast} $\cos(C, U)$: \textcolor{green!50!black}{correct} minus \textcolor{red!60!black}{wrong} branch, and helpful minus harmful $C$. \textbf{Nulls:} $C$ from another item at the same depth; the pre-divergence window (identical tokens, must read $0$).};
\end{tikzpicture}
\vspace{-2mm}
\caption{Decision-point design of \S\ref{sec:cotloops}: minimal pairs on both sides. (a) Programs that differ only in one extra pass over a block, graded helpful or harmful by the flip of the answer. (b) Chains sampled on the same item that share every token up to the decision point, so that question, prompt and reasoning so far cancel exactly in the contrast.}
\label{fig:schema}
\vspace{-4mm}
\end{figure}

A natural hypothesis is that a chain of thought is the model's way of re-applying its own layers to a refined state, so that a repairing cycle and a repairing chain would perform the same transformation. We tested this on the \emph{same problems} (GSM1k with DeepSeek-R1-Distill-Qwen-7B, MMLU with Qwen3.5-9B) with the design of Figure~\ref{fig:schema}: since a failed chain may contain the positive transformation and lose it downstream, every statistic is a contrast at the first token where two continuations of a \emph{split prefix} diverge.

\paragraph{Finding 4b: present, localised, and not what distinguishes a successful chain.} Table~\ref{tab:what}(b) gives the result. First, the trace of a helpful cycle \emph{is} in the chain: the best match with the chain's own layer updates lands at the cycle's own depth, far above the permuted-item null. Second, the perturbation is \emph{localised}: about ten effective dimensions, in largely the same coordinates across items. Third, none of it is about correction: harmful cycles leave the same trace as helpful ones, the trace is as strong in chains that fail as in chains that succeed, and the fix-minus-break and correct-minus-wrong directions are orthogonal at the branching token: cycles and chains are two different levers on the same parameters.

\section{Discussion, Limitations and Conclusion}
\label{sec:discussion}

\paragraph{Routing as a second axis of test-time adaptation.}
The versatility of a deployed model is not exhausted by the length of its chain of thought.
With no gradient update and one strategy for every model and task, a program of layers selected by the probability it assigns to a task's demonstration labels, and arbitrated by the model's own confidence, improves on the standard pass on average over MMLU, substantially on four of seven post-trained models, and most of all far from the training distribution: on ARC-AGI it doubles the accuracy of a 7B model whose \cotarm\ does not help (Tables~\ref{tab:mmlu} and~\ref{tab:arc}).
The two axes are complementary in a precise sense.
A program re-allocates depth within one forward pass, at a cost fixed by its length; a chain spends sequential tokens, at a cost bounded only by its budget.
The two succeed on different queries, their composition exceeds \cotarm\ alone (Table~\ref{tab:mmlu}), and the trace of a helpful route is no more present in chains that succeed than in chains that fail (Figure~\ref{fig:schema}).
Routing is thus not a cheaper form of reasoning but a different lever on the same parameters, and the two levers can be pulled together.

\paragraph{Where the potential is left untapped.}
The oracle over the three arms reaches $.816$ against $.692$ for \cotarm, and the gap is one of selection, not of existence: demonstrations rank programs almost perfectly in aggregate (Spearman $.97$ across the ARC-AGI pool) but say little about which query needs which program, and confidence can veto a switch but not navigate to the best arm.
Our analyses locate the loss in two places, each pointing at a different intervention.
The first is in \emph{surfacing} the potential.
The ability a fitted program exploits is present at a quarter of pre-training and grows through it ($+.05$ over the standard pass at 25\% of OLMo-3's pre-training, $+.06$ at its end), and a reproducible headroom of $+.06$ to $+.24$ persists at every stage of two training trajectories; what the first stage of post-training removes is the selectability, by saturating confidence and entrenching the default pass (Figure~\ref{fig:arc_cot}, Table~\ref{tab:ladder}).
The routing potential is therefore a property of the pre-trained representations that current post-training buries rather than erases; selection signals other than confidence, or post-training that keeps confidence informative about alternative computations, could recover it in existing models.
The second is in the \emph{robustness} of the model to the refinements routing introduces.
Helpful programs are localised, one extra pass over a short block whose output stays close to what the full model computes, whereas, matched on problem and cost, the programs that fail drive the residual stream out of the distribution the following layers expect (Tables~\ref{tab:what} and~\ref{tab:ood}).
Training that exposes layers to re-entered residual streams, as looped and layer-dropout training do for a fixed architecture~\citep{dehghani2019universal,fan2020layerdrop,geiping2025scaling}, would widen the set of programs that succeed, and with it the range over which a future model can adapt at test time without any training of its own.

\paragraph{Limitations.}
Fitting a program presupposes demonstrations of the test query's task, and the distance between the oracle and every selection rule is the open problem we leave.
Our claims concern one routing language (skips and cycles over contiguous blocks), 3--31B models and single-pass readouts on knowledge and abstraction tasks; the ARC-AGI result is on one 7B model, several baselines were budget-reduced (Appendix~\ref{app:baselines}), and the trajectory result rests on two ladders of small models.

\paragraph{Conclusion.}
Dynamic routing opens a second axis along which an existing model adapts to a new problem at test time: complementary to \cotarm, free of gradient updates, and exploitable today from a handful of demonstrations by a strategy that transfers across models and tasks.
Its gains are consistent, but its potential is larger, and the two places where selection loses it, the entrenchment of the default pass by post-training and the fragility of later layers to refined inputs, are properties of how models are trained rather than of routing itself.
We therefore see dynamic routing as a paradigm for extending the plasticity of LLMs in rapid test-time adaptation: of existing models, through better selection among the computations they already support, and of future ones, through training that keeps those computations selectable and robust.

\bibliography{iclr2027_conference}
\bibliographystyle{iclr2027_conference}

\appendix
\section{Execution of programs}
\label{app:exec}

Programs are executed by a per-family patch of the decoder that follows an arbitrary layer sequence. Repeated layers require a key--value cache indexed by program position: we allocate one cache slot per executed step, so that a layer applied twice attends to its own earlier keys and values from the corresponding position of the sequence, and the identity program is byte-identical to the vanilla model (verified per family with an identity bit-exactness smoke test before any number was trusted; a looped path was additionally verified to produce identical tokens with and without cache on the Qwen2 family). Families whose cache cannot be made position-indexed under the current \texttt{transformers} release (hybrid linear/full attention in Qwen3.5; the preallocated hybrid caches of Gemma-4, Ministral-3 and OLMo-3) run non-identity programs cache-free; for those, forced-choice scoring uses one forward per option (option texts teacher-forced after the answer scaffold, length-normalised) so that a flip relative to the identity path is an exact property of the program. Mixture-of-experts layers keep their own token router untouched; per-layer attention types (sliding/full, no positional encoding) travel with the layer wherever it executes. Scaffolds close a family's forced reasoning block (\texttt{</think>}, \texttt{[THINK][/THINK]}, \texttt{enable\_thinking=False}) before the answer scaffold; Ministral-3 receives a system prompt (its template injects thinking instructions otherwise), and Llama-3.1 a system prompt instructing it to copy the option text verbatim, which alone moves its direct accuracy from $.377$ to $.497$ and is used identically in every arm.

\section{Experimental details}
\label{app:setup}

\paragraph{MMLU.} 49 subjects, 20 test and 10 demonstration questions each, golds are option texts, answers normalised (whitespace, quotes, trailing period, case) before exact match. \dir/\pp: answer scaffold pre-filled, at most 120 new tokens, sampling per model card (e.g.\ $T{=}.6$, top-$p$ $.95$ for the R1 distillate; $T{=}.7$, top-$p$ $.8$, top-$k$ 20 for Qwen3; $T{=}1.0$, top-$p$ $.95$, top-$k$ 64 for Gemma-4), confidence recorded as in Eq.~\ref{eq:conf} together with its raw log-probability sum and token count. \cotarm: the same prompt without the scaffold, 10{,}000-token budget, answer parsed from the last balanced \texttt{\textbackslash boxed\{\}}, confidence by one extra forward that self-forces the emitted answer at the answer position (with \texttt{logits\_to\_keep} so that long chains on large vocabularies do not silently drop it). Fit items: each of the 10 demonstrations posed as the query with the other 9 in context, 16-token gold-evidence pass recording $p_{\mathrm{gold}}$ and greedy match.

\paragraph{Program search.} Per subject: 300 Monte-Carlo tree search rollouts (seed 13) with the reward of Eq.~\ref{eq:evidence}, $p_{\mathrm{skip}}{=}.12$, $p_{\mathrm{loop}}{=}.15$, blocks of at most 4 layers repeated at most 2 extra times, exploration $\epsilon{=}.2$; top-10 candidates kept; a rung or model enters the search only if its direct accuracy clears a floor ($.05$; $.35$ for Llama-3.1 whose weak regime otherwise admits degenerate programs). Two audit findings are on record: the search seeds were process-salted in early runs (irreproducible but unbiased; now deterministic), and an early version of the Ministral search ignored the system prompt, which was redone on the clean channel (identity best on 46/49 instead of 49/49; all conclusions unchanged).

\paragraph{Bagging.}
\label{app:bagging}
15 members, each fitted by its own 300-rollout search on 8 of the 10 demonstrations (a fixed design of 15 held-out pairs in which every demonstration is held out three times), weighted by the mean gold probability of the member's program on its held-out pair; \vote\ takes the weighted majority of the members' answers, soft-\vote\ additionally multiplies the weight by the member's confidence in its answer. The OLMo port reproduces the DeepSeek numbers ($.385/.407/.412$ for 5/15/soft-15 members) bit-for-bit on the original member arms. A \emph{pool-restricted} variant in which members select from one search's top-10 instead of searching was also run on every rung and never beat the standard pass.

\paragraph{Ablation: aggregation strategies.}
\label{app:ablation}
The vote strategies were run on the two models that bracket the capability range on which they could matter, DeepSeek-R1-Distill-Qwen-7B (weakest, most search structure) and Qwen3-8B (mid-range, whose single program loses 13 points), and are reported in Table~\ref{tab:ablation} rather than in Table~\ref{tab:mmlu}. Three observations. (i) On DeepSeek the bagged vote is the best deployable method by a wide margin, 12 points above chain-of-thought, and combining it with the reasoning arms by max-confidence keeps almost all of it. (ii) The gain is \emph{bagging diversity}: soft-voting the near-duplicate top-10 of a single search gives $.298$ on the same model, and the 15-member vote's members, fitted on different demonstration subsets, choose different programs. (iii) On Qwen3-8B the same vote lands below the standard pass (though 10 points above the single program), and folding it into the confidence switch costs a point against the switch over the single program: aggregation repairs a model whose default pass is weak and whose searched programs carry real structure, and has nothing to repair once the default pass is strong. On a 10-subject block of the DeepSeek bed the same members applied \emph{under} chain-of-thought compose with it (\dir\ $.227$, \cotarm\ $.404$, vote $.445$, vote under \cotarm\ $.524$).

\begin{table}[h]
\caption{Aggregation ablation on MMLU (macro accuracy) for the two models on which the 15-member bagged vote was run; the other rows repeat Table~\ref{tab:mmlu} for reference.}
\label{tab:ablation}
\begin{center}
\scriptsize
\begin{tabular}{lcc}
\toprule
Method & DeepSeek-R1 7B & Qwen3 8B \\
\midrule
\dir\ (standard pass)                              & .242 & .747 \\
\cotarm\ (10k tokens)                              & .294 & .724 \\
\pp\ (single fitted program)                       & .310 & .614 \\
soft-\vote$_{10}$ (one search's top-10)            & .298 & .659 \\
\vote$_{15}$ (bagged, hard majority)               & .407 & -- \\
soft-\vote$_{15}$ (bagged, confidence-weighted)    & \textbf{.412} & .719 \\
\cs\ \{\dir,\cotarm,\pp\}                          & .329 & \textbf{.750} \\
\cs\ \{\dir,\cotarm,\vote$_{15}$\}                 & .410 & .740 \\
\mrg\ (\cotarm\ default; \dir,\pp)                 & .327 & \textbf{.750} \\
\bottomrule
\end{tabular}
\end{center}
\end{table}

\paragraph{Selection variants and per-domain decomposition.}
\label{app:variants}
Table~\ref{tab:variants} lists every rule we evaluated on the seven models from banked logs. \cs: arg-max confidence with tie order \cotarm$\succ$\dir$\succ$\pp\ (an earlier scorer broke ties in favour of the correct arm; the effect was $\le.001$ and all reported numbers use the deployable tie-break). \mrg: default arm unless a challenger's confidence exceeds it by $\tau$, $\tau$ chosen by 5-fold cross-fitting over subjects from $\{0,.001,\dots,.5\}$, ties toward the lowest $\tau$. Rank-calibrated: per-arm confidence percentiles instead of raw confidences. Confidence-weighted vote: over the arms' answers, \cotarm\ on ties (best on Ministral, whose arm confidences are scale-mismatched). Demonstration-gated: per-subject switch \dir/\cotarm\ by the median demonstration fitness of the identity program against a cross-fitted threshold. Stable-candidate \pp: per subject, the candidate that wins the most leave-one-demonstration-out folds instead of the raw arg-max (it picks the arg-max on 46--49 of 49 subjects for every model: selection stability was never the failure mode). Per domain, pooled over the seven models, the \cotarm-default margin rule's gain over \cotarm\ is largest on high-school biology and physics ($+.107$ each), college computer science, high-school statistics and nutrition ($+.086$), and smallest on professional accounting ($-.107$), prehistory ($-.086$), abstract algebra and miscellaneous ($-.071$); the gains come from the \cotarm-weak models (Ministral, Qwen3, DeepSeek) on recall-flavoured subjects, the losses concentrate where \cotarm\ genuinely computes (Qwen3 $-.50$ on abstract algebra; professional accounting hurts on four models including both Gemmas, whose immediate-arm confidences there tie or exceed \cotarm's while being wrong). The Gemma columns are otherwise structural zeros: their cross-fitted $\tau$ says never to leave \cotarm. Confidence-scale mismatch on Llama: \dir\ $.87$, \pp\ $.79$, \cotarm\ $.59$ median.

\begin{table}[h]
\caption{All arms and selection rules on MMLU (macro accuracy; margins and thresholds are cross-fitted; ``$+$ on'' = models on which the rule beats always-\cotarm), the two oracles, and the number of subjects on which the search preferred the standard pass over every program.}
\label{tab:variants}
\begin{center}
\scriptsize
\setlength{\tabcolsep}{1.6pt}
\begin{tabular}{lcccccccccc}
\toprule
Rule & DeepSeek & Llama & Ministral & Qwen3 & Qwen3.5 & Gemma 26B & Gemma 31B & avg & $\Delta$\cotarm & $+$ on \\
\midrule
\mrg, \cotarm\ default & .327 & .625 & .627 & .750 & .841 & .893 & .922 & .712 & $+.019$ & 5/7 \\
rank-calibrated confidence         & .325 & .561 & .637 & .750 & .838 & .885 & .905 & .700 & $+.008$ & 4/7 \\
demonstration-gated regime & .292 & .626 & .618 & .734 & .823 & .887 & .909 & .698 & $+.006$ & 2/7 \\
\cs\ \{\dir,\cotarm,\pp\}             & .329 & .524 & .628 & .750 & .837 & .886 & .915 & .696 & $+.003$ & 4/7 \\
\mrg, \dir\ default, \cotarm/\pp\ & .332 & .518 & .627 & .741 & .833 & .886 & .915 & .693 & $+.001$ & -- \\
\cotarm\ (10k)                        & .294 & .629 & .554 & .724 & .823 & .895 & .929 & .692 & -- & -- \\
confidence-weighted vote           & .334 & .571 & .643 & .753 & .811 & .846 & .887 & .692 & $.000$ & 3/7 \\
2-arm \cs\ \{\dir,\cotarm\}           & .289 & .524 & .624 & .752 & .842 & .887 & .915 & .690 & $-.002$ & 3/7 \\
2-arm \cs\ \{\cotarm,\pp\}            & .342 & .509 & .606 & .692 & .836 & .886 & .917 & .684 & $-.009$ & 2/7 \\
\mrg, \dir\ default, \pp\ & .309 & .508 & .632 & .748 & .795 & .841 & .883 & .674 & $-.018$ & -- \\
\dir                               & .242 & .497 & .618 & .747 & .795 & .842 & .882 & .660 & $-.032$ & 2/7 \\
stable-candidate \pp               & .302 & .439 & .604 & .665 & .761 & .838 & .886 & .642 & $-.050$ & 2/7 \\
\pp\ (arg-max)                     & .310 & .438 & .604 & .614 & .766 & .838 & .886 & .637 & $-.056$ & 2/7 \\
soft-\vote$_{10}$ (one search's top-10) & .298 & .506 & .533 & .659 & .674 & .838 & .886 & .628 & $-.065$ & 1/7 \\
\midrule
\textit{\orc, per query (3 arms)}   & \textit{.530} & \textit{.774} & \textit{.783} & \textit{.855} & \textit{.901} & \textit{.928} & \textit{.943} & \textit{.816} & & \\
\textit{\orc, per subject (best arm)} & \textit{.390} & \textit{.644} & \textit{.677} & \textit{.792} & \textit{.853} & \textit{.906} & \textit{.936} & \textit{.742} & & \\
search prefers standard pass (subjects of 49) & 6 & 11 & 46 & 27 & 41 & 49 & 49 & & & \\
\bottomrule
\end{tabular}
\end{center}
\end{table}

\section{ARC-AGI beds}
\label{app:arc}

Tasks of ARC-AGI-1 are rendered as text (grids as nested Python lists, the prompt wording and colour legend of \citet{arcvl2025}), with the task's demonstrations as examples and the test input as query, and turned into a 4-way multiple choice by three distractor grids derived from the gold output. 262 of 400 training tasks and 177 of 400 evaluation tasks pass a 200-cell size filter; 1317/1317 variants round-trip to the original gold, an independent blind check solved 20/20 sampled items with 0 ambiguous or wrong golds; a tighter-distractor variant costs 2--4 points and changes no conclusion. Every item stores three selection orderings and three held-out orderings of the options. Forced choice is the arg-max over the four letter tokens after the answer scaffold, on which the model places $.999$--$1.000$ of its whole-vocabulary mass at every thinking budget, so the readout is the model's actual answer. \emph{Pool and verification:} the standard pass scores $.309$ on the training tasks; MCTS with 150 rollouts labelled $82/181$ of its misses as repaired, and re-scoring every discovered program on three \emph{held-out} orderings kept $38/181$ ($119/333$ ``valid'' programs survive vs $7/524$ of the rejected ones, matching the $1.6\%$ null); the 95-program pool is the union of 57 motif-designed and 39 verified searched programs (25 pure cycles, 31 cycle+skip programs, zero pure skips among the verified repairs). \emph{Fitting:} one-shot items (each demonstration posed as a query with the others in context) serve for fitting; the 3-demonstration format and the evaluation split are held out from program selection. With 1/2/3 demonstrations the selection accuracy on the unseen split is $.316/.328/.333$ at one ordering and $.424/.429/.429$ at eight; beyond about six measurements the curve is flat, so the gap to the existence bound is not estimation noise. Worst-case confidence across orderings predicts transfer better than mean confidence (wins 187/200 folds). Programs chosen per task: 23 distinct; the most frequent is a cycle over layers 16--18. \emph{Task-agnostic programs (not used in the main text):} fixed programs selected on the training split also transfer to the unseen split (best single program, cross-fitted over 200 splits, $.443$; a cycle over layers 15--18, $.446$; cycles over layers 1--4 and 16--17 with layer 23 skipped, $.475$), \emph{Chain-of-thought (Table~\ref{tab:arc}):} on 60 training tasks the standard pass with thinking and budget forcing~\citep{muennighoff2025s1} scores $.317/.350/.300/.200$ at 0/256/1{,}024/4{,}096 thinking tokens; every generation hits the cap and budget forcing fires on 92--100\% of them; per-ordering accuracy falls $.678/.600/.517$, so what collapses with budget is agreement across the three orderings rather than per-question knowledge; a 16{,}384-token arm on 27 evaluation tasks scored $.074$ (mean generation 14{,}274 tokens) before it was stopped for cost. A fixed cycle over layers 15--18 scores $.433/.467/.400/.333$ at the same budgets. \emph{Qwen3-8B on ARC-1} (176 tasks): $.733$ standard, $.699$ with searched-fitted programs and $.699$ with their vote; its chain-of-thought arm was not completed. An ARC-AGI-2 bed (62 public evaluation tasks) showed no repair for any of DeepSeek, Qwen3-8B or Qwen3.5-9B and is omitted.

\section{Trained baselines}
\label{app:baselines}

\paragraph{MindSkip / Router-Tuning.} Bias-free linear routers on the mean-pooled post-norm input of the attention sub-layer of the deepest $\lfloor N/2\rfloor$ layers except the last, zero-initialised, trained with a straight-through estimator on 5{,}000 samples of the Llama-Pro SFT mix (1 epoch, batch 64, seq.\ 2048, lr $10^{-5}$) with $\lambda\in\{.001,.01,.1\}$ and the paper's selection rule (realized capacity in $[.4,.6]$, lowest validation cross-entropy). Control arm with gating off reproduces the direct arm ($.744$ vs $.747$ on Qwen3-8B); a two-point dose response ($\lambda{=}.001\!\to\!.1$) confirms the loss is gating. Gemma-4 transfer arms realized capacities of $.77$--$.80$ because gradient checkpointing inerts the capacity loss at that scale; the few-shot arms were trained without checkpointing and are capacity-matched. The paper's own headline (Llama-3-8B $65.3\!\to\!65.1$ at $1.21\times$ speed) is measured by log-likelihood ranking of the options.

\paragraph{Dr.LLM.} Per-layer MLP routers (skip/execute/repeat) on windowed mean-pooled states, trained teacher-forced on length-aware MCTS labels ($\sim$4k samples per model from ARC-Easy/Challenge and DART-Math levels 1--5, the paper's search constants) with focal loss and effective-number class weights, evaluated with one routing decision per query frozen for decoding. Three properties of the released code (an interpolation that is a no-op for some families; per-step re-decision at decode; a hard-coded 28-layer cap) were resolved in favour of plain arg-max, single-pass routing. The transfer labels were generated for all five dense models (DeepSeek complete, 3{,}504 rows) but the routed transfer arm was deferred behind the few-shot branch.

\paragraph{\polar\ predictor.} The release predictor ($\sim$2.1M parameters over a frozen Qwen3-Embedding-0.6B encoder~\citep{qwen3embedding}; per-layer segmentation and skip/execute/repeat logits; beam-decoded to five candidate programs) trained with the release hyper-parameters on MCTS pools of DART-Math~\citep{tong2024dartmath} levels 1--5 generated for each of our models. Measured throughput forced a budget-reduced reproduction: 200 questions per level $\times$ 100 rollouts instead of the paper's $\sim$4{,}100 training questions and unpublished search budget, which is stated wherever its numbers appear. MMLU evaluation decodes the five candidates and reports pass@1; pass@5 ($.489/.694$ on DeepSeek/Llama) is an oracle over five programs and subject to the multiple-comparisons caveat of \S\ref{sec:oracle}.

\paragraph{Few-shot configurations.} Each router is additionally trained on our 490 fit items with the paper's hyper-parameters and selection rules otherwise unchanged: \polar\ and Dr.LLM on per-item valid/invalid splits of the subject's top-10 searched programs judged by greedy correctness on that item (items with no correct program dropped, matching both papers' solvable-only convention; for both Gemmas the only correct program is the identity), one predictor/router per model over all 49 subjects; MindSkip on a calibration set built from the fit prompts (demonstrations in context, gold text as the assistant turn), $\sim$77 optimizer steps as in the paper, $\lambda$ re-swept.

\begin{table}[h]
\caption{Trained depth-routing baselines on the same bed, in each paper's transfer configuration and trained on our 490 fit items (few-shot), against the standard pass and our direct-regime margin rule (no \cotarm\ anywhere). MindSkip rows carry realized attention capacity; $^\ddagger$ above the paper's $\sim$.5 target, i.e.\ milder gating; $^{*}$ partial arm; --: not run.}
\label{tab:baselines}
\begin{center}
\scriptsize
\setlength{\tabcolsep}{2.5pt}
\begin{tabular}{lcccccccc}
\toprule
Method & DeepSeek & Llama & Ministral & Qwen3 & Qwen3.5 & Gemma-26B & Gemma-31B & avg \\
\midrule
\dir\ (standard pass)             & .242 & .497 & .618 & .747 & .795 & .842 & .882 & .660 \\
\mrg\ (\dir\ default; \pp)        & \best{.309} & \best{.508} & \best{.632} & \best{.748} & \best{.795} & .841 & \best{.883} & \best{.674} \\
\midrule
MindSkip, transfer (capacity)     & .082 (.61) & .350 (.46) & .405 (.57) & .156 (.53) & .618 (.56) & .765 (.80)$^\ddagger$ & .740 (.77)$^\ddagger$ & .445 \\
MindSkip, few-shot                & .007 & .350 & .234 & .038 & .646 & -- & .838 & -- \\
Dr.LLM, few-shot                  & .249 & .349 & .621 & .687 & .786 & .789$^{*}$ & .882 & .623 \\
\polar\ predictor, transfer       & .231 & .383 & -- & -- & -- & -- & -- & -- \\
\polar\ predictor, few-shot       & .231 & .501 & .622 & .747 & .787 & .850$^{*}$ & .882 & .660 \\
\bottomrule
\end{tabular}
\end{center}
\vspace{-3mm}
\end{table}

\section{Training trajectories}
\label{app:ladder}
\begin{table}[h]
\caption{OLMo-3-7B trajectory on the MMLU bed (macro accuracy). Stage-1 percentages are of the pre-training token budget; ``base'' includes stages 2--3. s\vote$_{10}$: soft vote over one search's top-10; \vote$_{15}$/s\vote$_{15}$: (soft) vote of 15 members fitted on demonstration subsets; $\Delta$vote: best vote minus \dir. Bold: above the rung's standard pass by more than $.01$. ``conf.'': median answer confidence of the standard pass; ``identity'': subjects (of 49) on which the search preferred the standard pass. Oracles over \{\dir, top-10 programs\}: hindsight (first draw), \emph{transferable} (correct in both of two independent draws), luck (hindsight minus transferable), and the independence null (within-subject per-query shuffle, 20 repetitions).}
\label{tab:ladder}
\begin{center}
\tiny
\setlength{\tabcolsep}{1.5pt}
\begin{tabular}{lccccccccccccc}
\toprule
Rung & \dir & \pp & s\vote$_{10}$ & \cs & \mrg & \vote$_{15}$ & s\vote$_{15}$ & $\Delta$vote & conf. & identity & hinds. & transf. & luck / null \\
\midrule
stage-1, 5\%   & .299 & .293 & .276 & .299 & .294 & --   & --   & --      & $\sim$.6 & 46 & .733 & -- & -- / .870 \\
stage-1, 15\%  & .337 & .339 & --   & .345 & .338 & --   & --   & --      & $\sim$.6 & -- & .745 & -- & -- / .860 \\
stage-1, 25\%  & .392 & .376 & .387 & .381 & .386 & \best{.448} & \best{.448} & $+.057$ & .589 & 38 & .803 & .600 & $+.203$ / .936 \\
stage-1, 50\%  & .470 & .412 & .444 & .458 & .457 & \best{.478} & \best{.484} & $+.016$ & .597 & 23 & .833 & .672 & $+.160$ / .964 \\
stage-1, 100\% & .487 & .457 & .411 & .473 & .479 & \best{.531} & \best{.531} & $+.045$ & .582 & 42 & .855 & .665 & $+.190$ / .960 \\
base           & .533 & .518 & \best{.560} & .552 & .553 & \best{.580} & \best{.592} & $+.059$ & .749 & 19 & .866 & .768 & $+.098$ / .986 \\
SFT            & .602 & .564 & .569 & .603 & .597 & .595 & .599 & $-.003$ & .970 & 30 & .842 & .791 & $+.051$ / .983 \\
DPO            & .600 & .564 & .577 & .605 & .601 & .608 & .612 & $+.013$ & .984 & 33 & .831 & .794 & $+.037$ / .979 \\
RLVR (Instruct)& .601 & .524 & .562 & .608 & .608 & .590 & .595 & $-.006$ & .993 & 28 & .827 & .784 & $+.043$ / .974 \\
\bottomrule
\end{tabular}
\end{center}
\vspace{-3mm}
\end{table}

OLMo-3-7B rungs: \texttt{stage1-step71000} (5\%), \texttt{stage1-step212000} (15\%), \texttt{stage1-step350000} (25\%), \texttt{stage1-step700000} (50\%), \texttt{stage1-step1413814} (100\% of stage 1), \texttt{Olmo-3-1025-7B} (base), \texttt{Olmo-3-7B-Instruct-SFT}, \texttt{-DPO}, \texttt{Olmo-3-7B-Instruct} (RLVR); each post-training repo's base model chains back to the same base. All rungs share the instruct card's sampling ($.6/.95$) and the identical bed, scaffold text, evidence channel and search configuration as the seven main models; base checkpoints use a template-free few-shot scaffold.

\paragraph{Oracles along the ladder.} The hindsight oracle over \{\dir, top-10 candidates\} is $+.43$ to $+.23$ over \dir\ across the nine OLMo rungs (Table~\ref{tab:ladder}), but the union of independent answerers with the same marginal accuracies is higher at every rung, so the ten programs' successes are strongly correlated with each other and with \dir\ and contribute no complementary coverage beyond chance (hindsight minus null $-.11$ to $-.15$); the shrinkage of the apparent headroom with training tracks the shrinkage of the null. The \torc\ re-samples every arm in a second independent draw (card sampling, new seed) and counts a program for a query only if it is correct in both draws; for all rungs except SmolLM3 50\%, only the first-draw oracle winners were re-sampled (a lean retest, validated equal to the full 980-row retest on that rung). It is a conservative floor, since a good but stochastic program is penalised on both draws; the deployment ceiling lies between it and the hindsight oracle. Luck (hindsight minus transferable) is $+.16$ to $+.26$ on pre-training rungs and $+.04$ to $+.10$ after post-training, on both ladders. Across the OLMo rungs the per-query standard deviation of confidence across the ten candidate programs is $.14$--$.18$ everywhere and confidence declines monotonically with candidate rank at every rung, i.e.\ confidence ranks programs as the demonstrations do; the per-query correlation between the program's and the standard pass's confidence is $.1$--$.3$ on pre-training rungs and $.3$--$.5$ after alignment; the program's median confidence minus the standard pass's is $\approx 0$ on pre-training rungs, $+.028$ at the base model and $-.04$ at RLVR.

\paragraph{Selection rules along the ladder.} Besides the rows of Table~\ref{tab:ladder}, a pool-restricted 15-member vote (members select from one search's top-10 instead of searching) was run on every rung and never beat the standard pass; the canonical \cs\ and \mrg\ numbers use the same tie and $\tau$ semantics as the seven-model table (an earlier ladder scorer broke training ties toward the highest $\tau$, which inflated the aligned rungs by up to $.011$). Confidence arbitration is at or slightly above \dir\ only on the post-training rungs (up to $+.014$), where confidences are saturated but calibrated; on the pre-training rungs it is never positive beyond noise despite spread confidences, because the single program is worse than \dir\ and the programs carry no complementary signal.

\paragraph{The SmolLM3-3B ladder and the Apertus-8B rungs.} SmolLM3-3B rungs: stage-1 steps 240k/720k/1.16M/2.36M ($\approx$5/15/25/50\% of stage 1), stage-3 step 4.72M, the base model (raw few-shot scaffold throughout; the last two are transiently off-distribution for the boxed-answer format, which depresses their standard pass), and the SFT, APO-soup and final instruct checkpoints (chat scaffold). Table~\ref{tab:replication} lists every rung with the same oracle columns as Table~\ref{tab:ladder}; the bagged vote was not run on this ladder. The pattern is the OLMo one: the single fitted program never beats the standard pass by more than noise (it exceeds it only on the two raw-scaffold checkpoints, where the standard pass is depressed), the hindsight oracle sits below the independence null at every rung, luck shrinks from $\approx+.22$ to $\approx+.10$ with post-training, and the transferable oracle stays $+.06$ to $+.19$ above the standard pass throughout. Search structure differs by family: SmolLM3 finds non-identity winners on a third of subjects at every completed rung (identity preferred on 43/32/32/30 of 49 at 5/15/25/50\%), OLMo follows an inverted U, and Apertus-8B, whose ladder is paused after three rungs (5.7/15.2/24.7\% of its 15T tokens: standard pass $.300/.317/.340$, fitted program $.259/.318/.311$, identity preferred on 49/48/48 subjects), is identity-dominated from the start. Median answer confidence is spread on the base checkpoints ($.59$ Apertus, $.64$ SmolLM3) and saturated on the instruct checkpoints ($.90/.91$).

\begin{table}[h]
\caption{SmolLM3-3B trajectory on the MMLU bed (macro accuracy). \pp: single fitted program; oracles over \{\dir, top-10 programs\} as in Table~\ref{tab:ladder}: hindsight (first draw), transferable (correct in both draws), luck (hindsight minus transferable), independence null; ``identity'': subjects (of 49) on which the search preferred the standard pass, where recorded. $^{r}$: raw few-shot scaffold with depressed answer-format compliance.}
\label{tab:replication}
\begin{center}
\scriptsize
\setlength{\tabcolsep}{5pt}
\begin{tabular}{lcccccccc}
\toprule
Rung & \dir & \pp & hindsight & transferable & luck & null & transferable $-$ \dir & identity \\
\midrule
stage-1, 5\%   & .252 & .224 & .625 & .370 & $+.255$ & .754 & $+.118$ & 43 \\
stage-1, 15\%  & .262 & .242 & .675 & .439 & $+.237$ & .843 & $+.176$ & 32 \\
stage-1, 25\%  & .332 & .312 & .743 & .518 & $+.225$ & .920 & $+.187$ & 32 \\
stage-1, 50\%  & .362 & .310 & .740 & .525 & $+.214$ & .939 & $+.163$ & 30 \\
stage 3$^{r}$  & .263 & .266 & .660 & .456 & $+.204$ & .793 & $+.193$ & -- \\
base$^{r}$     & .204 & .236 & .582 & .366 & $+.215$ & .701 & $+.162$ & -- \\
SFT            & .484 & .479 & .652 & .548 & $+.104$ & .810 & $+.064$ & -- \\
APO soup       & .556 & .523 & .740 & .653 & $+.087$ & .892 & $+.097$ & -- \\
final (Instruct) & .549 & .551 & .765 & .661 & $+.104$ & .913 & $+.112$ & -- \\
\bottomrule
\end{tabular}
\end{center}
\vspace{-3mm}
\end{table}

\section{BIG-bench and mathematics}
\label{app:bb}

On 113 BIG-bench tasks~\citep{srivastava2023bigbench} (81 multiple-choice, 32 generative; 10 demonstrations, 4 held-out queries per task) with DeepSeek-R1-Distill-Qwen-7B, the per-task fitted program from a fixed 96-program pool scores $.518$ against $.425$ for the standard pass ($p{=}1.2{\times}10^{-6}$), $.456$ for the best constant program, $.425$ for a derangement null that routes each task by another task's evidence, and $.706$ for the pool oracle; $.438$ for 512-token and $.507$ for 2{,}048-token \cotarm, the latter statistically indistinguishable from the program ($p{=}.91$) and winning on different tasks (38/36 split). On the 250 \texttt{list\_functions} sub-tasks the program scores $.421$ against $.368$ standard and $.136/.311$ for 512/2{,}048-token \cotarm\ (98\%/77\% of chains budget-forced). On the 20-task fusion bed (up to 50 queries per task): standard $.457$, \cotarm\ $.571$, 15-member vote $.523$, the same members \emph{under} \cotarm\ $.616$, a demonstration-gated switch among the regimes $.678$, per-query confidence switch $.628$. The 20-task bed was selected adversarially for the 7B model; on it, Gemma-4-26B ($.902$ standard) and Qwen3-8B ($.785$) gain nothing from programs transplanted from the 7B's pool (this pool-transplant setting was superseded by model-specific search everywhere else in the paper).

\label{app:negative}
On GSM8K and GSM1k~\citep{cobbe2021gsm8k,zhang2024gsm1k} with same-dataset demonstrations and \emph{un-normalised} gold log-likelihood as fitness, per-problem selection falls below the standard pass ($.107$ vs $.160$, $.075$ vs $.105$) and below the derangement null: the full model ranks 37--41 of 96 in demonstration fitness despite being the best decode program, because cycles sharpen the output distribution and inflate the absolute gold mass without improving the arg-max. The multiple-choice fitness of ARC and MMLU is normalised over a contrast set, which cancels this; free-form fitness needs one too. Enlarging the candidate pool from 5 to 95 programs lowers the selected accuracy while raising the oracle (winner's curse). A learned generator of programs (LoRA on hidden states, trained on up to 106k valid MCTS transitions over 7{,}473 GSM8K problems) never beat the standard pass, collapsing to one good constant program; a linear scorer over 31 label-free dynamics features of a candidate's trajectory reaches AUC $.85$ for validity and, used as a router with the standard pass as a competing candidate, gains $+3$ of 120 held-out problems.

\section{Analyses: statistics and nulls}
\label{app:cotloops}

\paragraph{Helpful vs.\ harmful cycles (\S\ref{sec:what}, Table~\ref{tab:what}).} The GSM8K direct-answer search runs the MCTS of \S\ref{sec:fewshot} with the answer in the loop over the 7{,}473 training problems (full-depth accuracy $24.5\%$, 61\% of misses repairable, oracle $70\%$, 106{,}129 valid transitions; on the first 100 problems, 634 valid non-full programs of which 505 mix, 123 loop, 6 pure skip). Matched pairs are mined from this supervision for DeepSeek-R1-Distill-Qwen-7B: 12{,}444 forward pairs (cycle fixes: program correct, same program without the cycle wrong) over 2{,}272 problems and 14{,}261 reverse pairs (cycle breaks) over 1{,}522 problems, nearly all pure-loop, from which 500+500 are sampled and their trajectories captured prefill-only. The fixed-cycle scan executes every single-block cycle of width $\le 4$ with 1--4 extra passes on GSM8K and averages accuracy over positions ($.258/.192/.115/.070$ vs $.300$). On the MMLU fits the median block start is layer 10 of 32 on Qwen3.5 with width $\approx 2.5$. Features: participation ratio of the trajectory's step deltas, block contraction $d_2/d_1$, norm ratio, curvature, final cosine; an eight-pass forced orbit of the block (spectral radius, settling residual, attractor class; the state moves less at the second application than at the first, then expands, norm $\times 5$ over eight passes); Mahalanobis, ViM and $k$NN distances to normal layer-$L$ states (Mahalanobis $\approx 59$ for both fixing and breaking cycles vs $\approx 15$ for normal states); the update magnitude of the layer after the block relative to its usual magnitude; and the label-free logit-lens confidence of the final state (max-probability, margin, entropy) as the cycle's \emph{change} of it. A classifier on heterogeneous (skip-carrying) counterfactual pairs collapses to chance on every geometric family; only the counterfactual confidence shift and the per-problem-cleansed representation difference reach $.58$--$.65$. The 31-feature validity scorer (Kullback--Leibler divergence and top-1 agreement of the logit-lens output distributions with the full model, displacement, norm ratio, contraction, removed contribution, spikiness, program shape, a 7-bin depth profile of trajectory divergence) is fitted on 6{,}828 train and 2{,}221 test rows split by problem; linear AUC $.804$, MLP $.758$, positive-unlabelled $.805$; ablating any family costs $\le.002$; its dominant coordinate is the Kullback--Leibler proximity to the full model's output. The within-problem repair probe is ridge regression on 16 principal components over a problem's cycles exiting within $\pm4$ layers of $L$, 5-fold cross-validated per problem (median AUC $.759$; $.508$ permuted; $.612$ from block position; $.527$ from $\|C\|$; $.544$ trained on layers 0--15 and tested on 16--31; $.55$ across problems). Direction statistics: cosine of the cycle's perturbation with the block's own first-pass update $+.65$ (vs $+.09$ for a different block); cosine between the fixing and breaking class means $+.98$. ARC motif statistics contrast the 56 verified repairs with the search-evaluated programs that failed verification on the same tasks. \emph{Out-of-distribution test (Table~\ref{tab:ood}):} from the DART-Math search pools of DeepSeek-R1-Distill-Qwen-7B, one pair per problem (419 pairs) of programs that both contain a cycle and have identical total and repeated pass counts, one correct and one wrong, so that loop presence and compute are held fixed; each program's final (layer-27) state, the input to the final norm and head, is scored against the full-depth reference cloud in a per-layer 256-dimensional principal subspace by Mahalanobis and cosine distance. Paired wrong-minus-correct differences: cosine $+.069$ [$+.050,+.081$], wrong further in 69\% of pairs, AUC $.654$; Mahalanobis $+.46$ [$-.07,+1.13$], AUC $.554$. Raw fix/break pairs (loop vs.\ no loop) show only a loop-presence artifact, and unpaired loop-on states give AUC $.52$; the effect appears only in the paired, count-matched, direction-only comparison. Mid-network per-layer trajectories and an MMLU replication are pending.

\paragraph{Chain-of-thought vs.\ cycles (\S\ref{sec:cotloops}, Figure~\ref{fig:schema}, Table~\ref{tab:what}(b)).} Beds: GSM1k with DeepSeek-R1-Distill-Qwen-7B (74 mixed problems, 481 fixing vs 589 non-fixing chains, $k{=}32$ samples; cycles from MCTS supervision) and MMLU with Qwen3.5-9B (95 items with 122 single-block programs each from the exhaustive scan; 380 chains for whole-chain contrasts; 213 split prefixes over 32 items, 1{,}961 continuations, for decision-point contrasts). Targets: $C=h_{\mathrm{on}}(b_1)-h_{\mathrm{off}}(b_1)$ per (item, cycle), graded helpful/harmful by the change of the gold margin and matched within item ($K{=}5$ per item, 475 vs 475); the within-item ridge discriminant on 16 principal components over cycles exiting within $\pm 4$ layers of $L$; descendants $C(L)=\mathrm{traj}_{\mathrm{loop}}[L{+}w]-\mathrm{traj}_{\mathrm{id}}[L]$; the same cycle on the \cotarm\ prompt. Candidates: $U(t)=A[L{+}1,t]-A[L{+}1{-}k,t]$ at anchors $t$ at or after the token where the two branches of a split prefix first differ (an absolute offset ladder of 17 anchors), $k\in\{1,\dots,4\}$. Statistics: max-cosine over anchors and widths, the same at the cycle's own depth, concentration of the arg-max offset at the cycle's own block, counted presence against the permuted null's percentile, principal angles between item-wise program and chain subspaces, cosine of fix-minus-break and correct-minus-wrong axes, and held-out $R^2$ of a ridge map in an item-split principal-component basis against the identity map and a shuffled pairing. Dimensionality: over 1{,}902 fixing cycles the perturbations have $\approx 9.5$ effective dimensions (participation ratio; one direction carries 29\% of the variance, five carry 55\%); top-64 coordinate sets overlap 20.8/64 between problems ($20\times$ chance). Mandatory preprocessing: queries residualised against the top-8 shared principal components fitted out of fold (raw queries have participation ratio 4.4 of 468 and mean pairwise cosine $+.34$). Nulls: target from a different problem at the same depth; spectrum-matched random query; the pre-divergence window (always $\approx 0$); the shared-anchor control on the hybrid-attention model, whose chunked linear-attention states are numerically length-dependent. Two artifacts that would have produced false positives are on record: padding of batched continuations (anchors placed inside padding; offset-0 excess $+.083\!\to\!+.020$ once stripped) and a massive-activation coordinate that carries 29\% of the state's energy at layer 20 and inflated every state cosine about five-fold (dropping it takes the best match between a fixed state and the chain's state from $.83$ to $.17$). Aggregation is within item, bootstrapped over items; a best-of-$N$ statistic is compared only with an identically searched null.

\end{document}